\documentclass[10pt,twocolumn,letterpaper]{article}

\usepackage[pagenumbers]{cvpr} 

\usepackage{graphicx}
\usepackage{amsmath}
\usepackage{amssymb}
\usepackage{booktabs}

\usepackage[pagebackref,breaklinks,colorlinks]{hyperref}

\newcommand*{\affaddr}[1]{#1} 
\newcommand*{\affmark}[1][*]{\textsuperscript{#1}}
\renewcommand\thefootnote{}

\newcommand{\del}[1]{}

\usepackage[capitalize]{cleveref}
\crefname{section}{Sec.}{Secs.}
\Crefname{section}{Section}{Sections}
\Crefname{table}{Table}{Tables}
\crefname{table}{Tab.}{Tabs.}

\def\cvprPaperID{5065} 
\def\confName{CVPR}
\def\confYear{2023}

\begin{document}

\title{Leveraging Inlier Correspondences Proportion for Point Cloud Registration}


\author{
Lifa Zhu \textsuperscript{$\dag$}\affmark[1] \qquad Haining Guan \textsuperscript{$\dag$}\affmark[2] \qquad Changwei Lin \affmark[1] \qquad Renmin Han \textsuperscript{*}\affmark[3]\\
\affaddr{\affmark[1]DeepGlint} \qquad
\affaddr{\affmark[2]Beihang University} \qquad \affaddr{\affmark[3]Shandong University} \\
{\tt\small zhulf0804@gmail.com, hnguan@buaa.edu.cn, changweilin@deepglint.com, hanrenmin@sdu.edu.cn} 
}

\maketitle

\begin{abstract}
\vspace{-.5em}
In feature-learning based point cloud registration, the correct correspondence construction is vital for the subsequent transformation estimation. However, it is still a challenge to extract discriminative features from point cloud, especially when the input is partial and composed by indistinguishable surfaces (planes, smooth surfaces, etc.). As a result, the proportion of inlier correspondences that precisely match points between two unaligned point clouds is beyond satisfaction. Motivated by this, we devise several techniques to promote feature-learning based point cloud registration performance by leveraging inlier correspondences proportion: a pyramid hierarchy decoder to characterize point features in multiple scales, a consistent voting strategy to maintain consistent correspondences and a geometry guided encoding module to take geometric characteristics into consideration.
Based on the above techniques, We build our \textbf{G}eometry-guided \textbf{C}onsistent \textbf{Net}work (GCNet), and 
challenge GCNet by indoor, outdoor and object-centric synthetic datasets. Comprehensive experiments demonstrate that GCNet outperforms the state-of-the-art methods and the techniques used in GCNet is model-agnostic, which could be easily migrated to other feature-based deep learning or traditional registration methods, and dramatically improve the performance. 
   \footnote{\textsuperscript{$\dag$} equal contributions and \textsuperscript{*} corresponding author.}
   \setcounter{footnote}{0}
   \renewcommand\thefootnote{\arabic{footnote}}
\end{abstract}

\vspace{-1.5em}
\section{Introduction}
\vspace{-.25em}
\label{sec:intro}



Point cloud registration tries to find a transformation that best aligns two overlapping point cloud fragments, serving as a key component in various downstream tasks such as 3D reconstruction\cite{geiger2011stereoscan} and SLAM\cite{salas2013slam++,zhang2015visual}. In the past decades, various methods have been proposed to solve the point cloud registration problem, such as methods based on direct transformation optimization \cite{besl1992method,rusinkiewicz2001efficient,yang2015go} and hand-crafted descriptors\cite{rusu2008aligning,rusu2009fast,tombari2010unique}. When it comes to the deep learning era, two categories of methods have been developed: {\bf end-to-end} learning methods and {\bf feature-learning} based methods. The end-to-end learning methods achieve both point feature learning and transformation estimation in one forward pass, some of which \cite{wang2019deep,wang2019prnet,yew2020rpm,fu2021robust,choy2020deep,lee2021deep,lu2021hregnet,qin2022geometric} rely on correspondences establishment for subsequent Procrustes analysis, while the others pay more attention to global features between the point clouds\cite{aoki2019pointnetlk,sarode2019pcrnet,huang2020feature,xu2021omnet,xu2021finet}. The feature-learning based  methods\cite{rusu2008aligning,rusu2009fast,tombari2010unique,deng2018ppfnet,deng2018ppf,gojcic2019perfect,choy2019fully,bai2020d3feat,huang2021predator,ao2021spinnet,yu2021cofinet} concentrate on the extraction of geometric and other useful information of the point clouds to compose discriminative local features, to build correspondences based on nearest neighbor search in feature space and estimating the 3D rigid transformation by robust pose estimators\cite{fischler1981random,yang2020teaser}. 
Here, this paper focus on the discussion of feature-based point cloud registration methods.

Feature-learning based registration methods originated from hand-crafted descriptors, such as FPFH\cite{rusu2008aligning,rusu2009fast} and SHOT\cite{tombari2010unique}. These hand-crafted descriptors are devised to discover the local invariance in point clouds and find the pairwise correspondences that assist transformation estimation. However, to design an efficient and robust descriptor that fully utilizes the point information is non-trivial. Recently, great progress has been made 
with deep learning\cite{deng2018ppfnet,deng2018ppf,gojcic2019perfect,choy2019fully,bai2020d3feat,huang2021predator,ao2021spinnet,yu2021cofinet}. FCGF\cite{choy2019fully} extracts point features by utilizing a sparse 3D fully-convolutional network with metric learning to extend receptive field. D3Feat\cite{bai2020d3feat} utilizes a joint learning of feature detector and descriptor, demonstrating the superiority over other methods,  especially when using a small number of keypoints. PREDATOR\cite{huang2021predator} learns point attributes, such as overlap and matchability, to help build correspondences more precisely. 

\begin{figure}[!t]
\centering
\vspace{-0.7em}
\includegraphics[width=.48\textwidth]{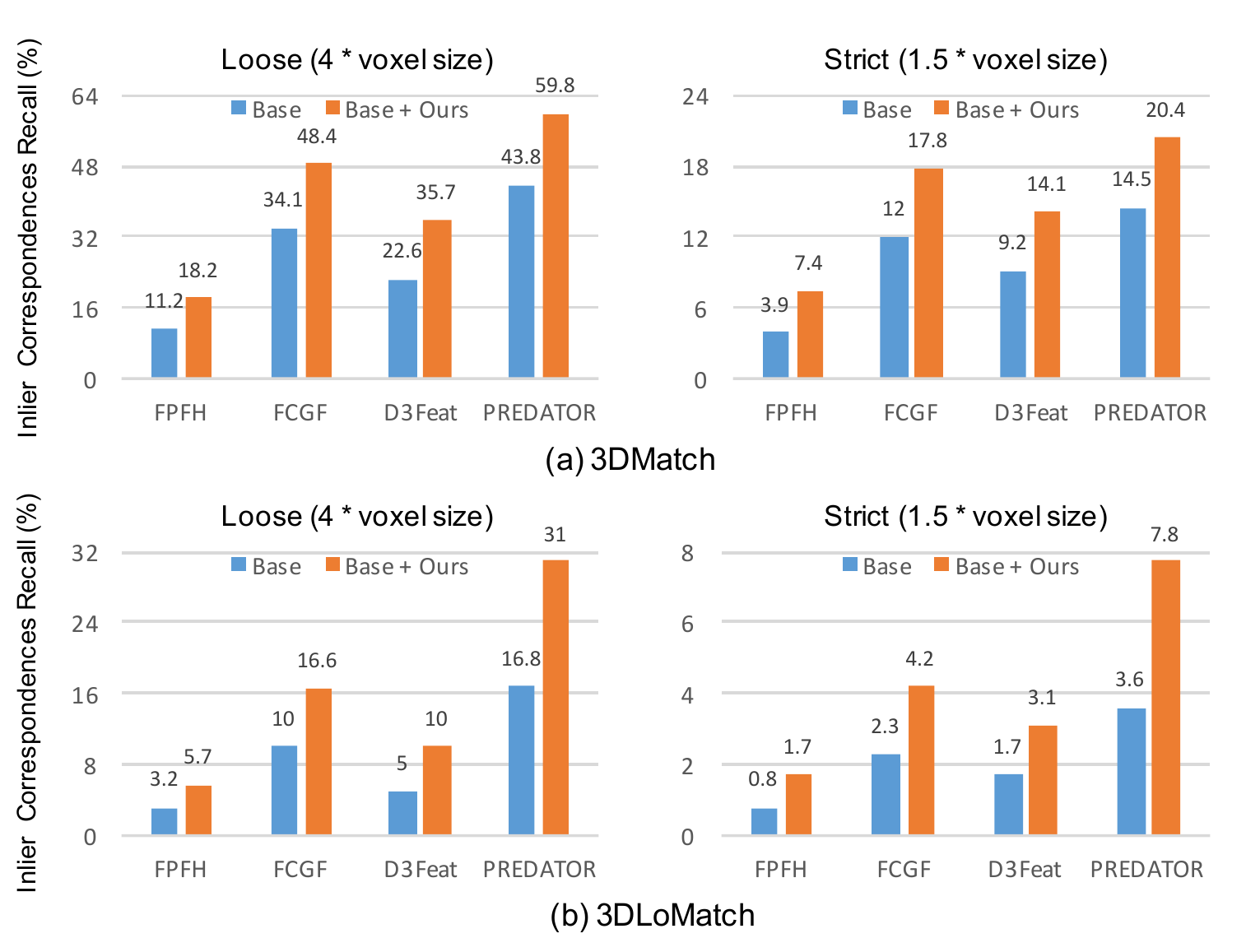}
\vspace{-2.5em}
\caption{Low inlier correspondences recall on (a) 3DMatch and (b) 3DLoMatch across traditional and learning-based methods. Two different thresholds for inlier correspondences are considered. With our proposed techniques, the inlier correspondences proportion among the putative correspondences is promoted stably.} 
\label{fig:intro}
\vspace{-1.75em}
\end{figure}

Though the feature-based registration methods have been widely studied, it's still challenging to obtain satisfactory inlier correspondences  through nearest neighbor search in feature space.
We investigate the inlier correspondence recall on both 3DMatch and it's low-overlap version 3DLoMatch datasets to validate the claim, as shown in \cref{fig:intro}. It turns out that defacto inlier correspondences recall among traditional and popular deep learning methods is relatively low, especially on the more challenging 3DLoMatch benchmark, resulting in a tough pose estimation. 
Thus, designing transferable strategies to promote the inlier proportion among the putative correspondences is  inevitable for robust and accurate point cloud registration.

Motivated by this, to abandon points with non-robust features (to produce fewer false matches) and to enhance the point geometric features (to recall more correct matches) are two effective ways that helps lift the proportion of correct correspondences. Based on the above discussions, we first proposed features-guided consistent voting mechanism accompanied with pyramid hierarchical architecture to reject points with specious features. Then we proposed a geometric-guided encoding module to maximally utilize the geometric information for accurate point matching in feature space. Finally, based on the encoder-decoder architecture, we construct Geometric-guided Consistent Network (GCNet) for robust and accurate point cloud registration. Experiments on 3DMatch and 3DLoMatch demonstrate that the proposed techniques are model-agnostic, able to be easily migrated to other networks or traditional methods, boosting their performance stably. Furthermore, GCNet achieved the state-of-the-art on indoor, outdoor and object-centric synthetic datasets. Particularly, GCNet achieves 92.9\% and 71.9\% Registration Recall on 3DMatch and 3DLoMatch, respectively.

In summary, our main contributions are as follows:
\vspace{-0.5em}
\begin{itemize}
\setlength{\itemsep}{0.25pt}
\setlength{\parsep}{0pt}
\setlength{\parskip}{0pt}
\item A consistent voting mechanism coupled with pyramid hierarchy architecture and a geometry-guided encoding module are proposed to lift the inlier correspondence proportion.
\item The 
GCNet outperforms the-state-of-art methods on indoor, outdoor and object-centric synthetic datasets.
\item The proposed techniques are model-agnostic, able to be easily migrated to other networks (or traditional methods) and stably improve the performance.
\end{itemize}

\vspace{-1.25em}
\section{Related Work}
\vspace{-.25em}
\subsection{End-to-end learning registration methods}
Recently, several 
progresses have been made in end-to-end learning methods. 
Some works \cite{aoki2019pointnetlk,sarode2019pcrnet,huang2020feature,xu2021omnet,xu2021finet} develop correspondence-free models based on global features for registration. Others\cite{wang2019deep,wang2019prnet,yew2020rpm,fu2021robust,zhu2021point} extract point features, establish correspondences, and obtain the transformation based on a differentiable SVD module. Until recently, end-to-end methods have shifted gradually to large-scale real-world datasets.  DHVR\cite{lee2021deep}, RegTR\cite{yew2022regtr} and  GeoTransformer\cite{qin2022geometric} achieve great performance on indoor and outdoor datasets by identifying the transformation parameter 
that best represents the consensus
, directly predicting the final set of correspondences, and using geometric Transformer to learn geometric features to match, respectively. 

\subsection{Feature-learning based registration methods}

Different point cloud encoding network for feature-learning have been proposed in recent years. 
PPFNet\cite{deng2018ppfnet} and PPF-FoldNet\cite{deng2018ppf}
introduces hand-crafted PPF\cite{drost2010model} for feature encoding in a patch-wise way. FCGF\cite{choy2019fully} adopts a fully-convolutional 3D network for dense feature extraction in one-forward-pass. 3DSmoothNet\cite{gojcic2019perfect} adopts smoothed density representations for point cloud representation. 
SpinNet\cite{ao2021spinnet} proposes spatial cyclindrical convolutions to convert local patches into rotation-invariant features. MS-SVConv\cite{horache20213d} consumes multi-scale point clouds with different downsampling 
for multiple features extraction.

Furthermore, some strategies are explored to boost the feature-learning based registration performance. D3Feat\cite{bai2020d3feat} and  PREDATOR\cite{huang2021predator} learn both point attributes (such as saliency and overlap) and point features for registration. CoFiNet\cite{yu2021cofinet} leverage a coarse-to-fine strategy to build correspondences more accurately. However, though large improvements have been made in feature-based methods, the inlier correspondences ratio is relatively low. 

\begin{figure*}[!t]
\centering
\vspace{-.5em}
\includegraphics[width=.81\textwidth]{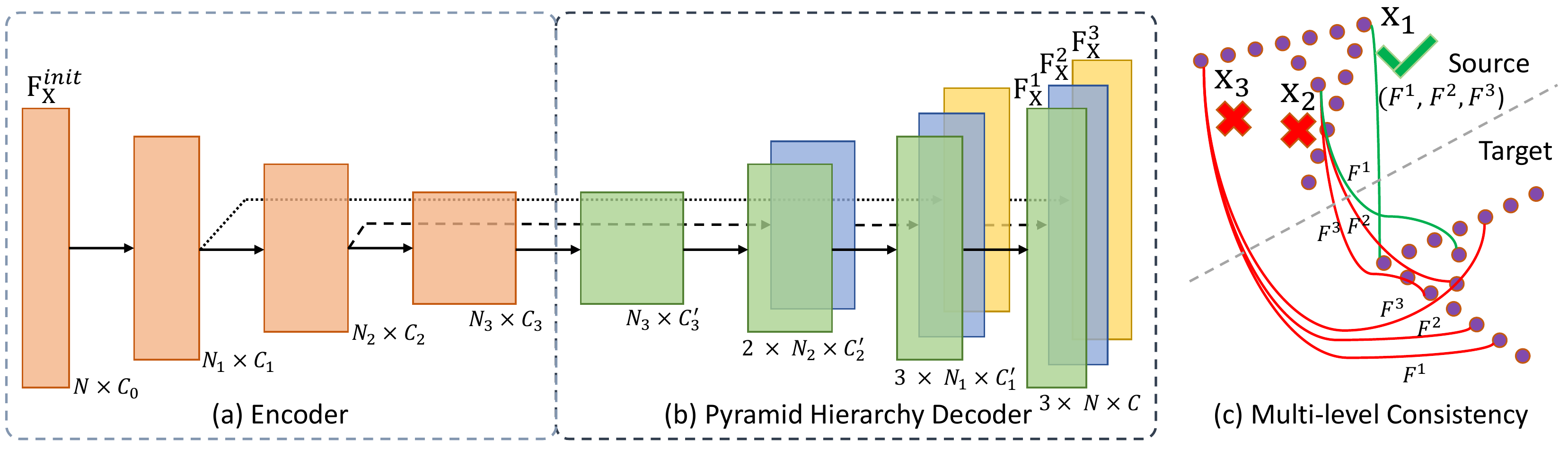}
\vspace{-.75em}
\caption{Pyramid hierarchy decoder and multi-level consistency for voting. Inputted 3D coordinates and skip connections are omitted in (a) and (b) for simplification. 
Green and red curves in (c) represents inlier and outlier correspondences, respectively.}
\vspace{-1.5em}
\label{fig:consistency}
\end{figure*}

\vspace{-.25em}
\subsection{Pose estimators}

Once putative correspondences are obtained in feature-based registration, pose estimators\cite{fischler1981random,leordeanu2005spectral,yang2020teaser,yang2019performance,choy2020deep,choy2020high} are utilized to calculate the final transformation. RANSAC\cite{fischler1981random} iteratively selects several points, calculates and evaluates transformation. TEASER\cite{yang2020teaser} formulates registration as a Truncated Least Squares (TLS) cost and decouples scale, rotation, and translation estimation based on graph-theory. Learning-based DGR\cite{choy2020deep} utilizes 6D sparse convolutional network\cite{choy2020high} to estimate the correspondence likelihoods used as weights for Procrustes analysis. PointDSC\cite{bai2021pointdsc} extends DGR using spatial consistency and spectral matching to estimates the correspondence inlier confidence. In this work, we choose the widely-used RANSAC for pose estimation.

\vspace{-.25em}
\section{Method}
\vspace{-.25em}
Given the source point cloud $\mathcal{X}=\{\mathbf x_i \in \mathbb R^3\}_{i = 1, 2, \cdots, N}$ and target point cloud $\mathcal{Y}=\{\mathbf y_j \in \mathbb R^3\}_{\ j = 1, 2, \cdots, M}$, where $N$ and $M$ are the point number in $\mathcal{X}$ and $\mathcal{Y}$, the aim of point cloud registration is to find a transformation $T \in \mathbf{SE}(3)$ that best aligns $\mathcal X$ and $\mathcal Y$, that is: 
\begin{equation}
\arg \min_{T}\frac{1}{|\sigma|}\sum_{(i, j) \in \sigma}||T(\mathbf x_i) - \mathbf y_{j}||_2,
\end{equation}
\vspace{-.25em}
where $\sigma$ denotes the definite yet unknown correspondence set and $|\cdot|$ denotes the cardinal number. For the convenience of the following discussion, all the data objects are denoted by matrix but not set.

Without loss of generality, we utilize source point cloud as an example to deliberate the following contents if source and target point cloud are not involved at the same time. Generally, inputted the source point cloud $\mathbf X \in \mathbb R^{N \times 3}$, the network outputs features $\mathbf F_{\mathbf X} \in \mathbb R^{N \times C}$, with each row corresponding to $C$-dimensional feature of a point.

%
\vspace{-.25em}
\subsection{Consistent correspondence}
\vspace{-.25em}

Partially overlapping and indistinguishable structures (smooth surfaces, virtual lines, etc) in point clouds make it challenging to extract discriminative features, thus generating relative low ratio of inlier correspondences for registration. Here, we propose to abate the proportion of specious correspondences with pyramid hierarchy decoder and multi-level consistency voting mechanism.

\textbf{Pyramid hierarchy decoder. }
The encoder-decoder network is usually used in feature-learning based methods. Inspired by BAAF-Net\cite{qiu2021semantic}, which gradually explores the point cloud in decreasing resolution for semantic segmentation, we introduce the pyramid hierarchy decoder for point cloud registration (\cref{fig:consistency}). 

Given source point cloud $\mathbf{X}$ with initial descriptor $\mathbf F_{\mathbf X}^{init}$ (generally initialized to the matrix of ones), the encoder process $\mathbf{X}$ with $L$ sub-sampling layers, producing multi-resolution point clouds $\left\{\mathbf{X}_1, \cdots, \mathbf{X}_L\right\}$ and features $\left\{\mathbf F^{en}_{\mathbf X_1}, \cdots, \mathbf F^{en}_{\mathbf X_L}\right\}$. Pyramid hierarchy decoder progressively upsamples the multiple resolution point clouds. Finally, it learns $L$-levels full-sized point cloud features $\left\{\mathbf F^1_{\mathbf X}, \mathbf F^2_{\mathbf X}, \cdots, \mathbf F^L_{\mathbf X}\right\}$. For $l$-th level feature $\mathbf{F}^l_{\mathbf{X}}$, it first upsamples $\mathbf{X}_{L+1-l}$ via 
\begin{equation}
    \text{MLP}(\text{cat}[\text{Up}(\text{MLP}(\mathbf{F}^{en}_{\mathbf{X}_{L+1-l}})), \mathbf{F}^{en}_{\mathbf{X}_{L-l}}]),
\end{equation}
where $\text{Up}(\cdot)$ is the nearest upsampling and cat$[\cdot,\cdot]$ is the concatenation operation. Then similar upsampling operations are conducted 
until obtaining the full-sized point cloud. Schematic diagram of pyramid hierarchy decoder for $L$$=$$3$ is shown in \cref{fig:consistency}(b). Each level's feature starts being upsampled from point cloud with different resolution, thus the point receptive field for each level's feature is different.

\textbf{Consistent voting. }
Inspired by Scale Invariant Feature Transform(SIFT)\cite{lowe1999object,lowe2004distinctive}, which locates and describes high contrast keypoints based on scale space, we propose multi-level consistency voting mechanism based on multi-scale features that rejects correspondences with low confidence.

The voting mechanism is feature-guided to get the robust correspondences. The consistent voting accepts $\left\{\mathbf{X}, \mathbf F^1_{\mathbf X}, \mathbf F^2_{\mathbf X}, \cdots, \mathbf F^L_{\mathbf X}\right\}$ and $\left\{\mathbf{Y}, \mathbf F^1_{\mathbf Y}, \mathbf F^2_{\mathbf Y}, \cdots, \mathbf F^L_{\mathbf Y}\right\}$ as input, and outputs the final consistent correspondences set $\mathcal{C}=\left\{(\mathbf{x}_i, \mathbf{y}_{\sigma(i)}) \,| \, i = 1, 2, \cdots, N \right\}$, where $\sigma(i)$ is $\mathbf{x}_i$' correspondence point index in $\mathbf{Y}$ or -1 denoting an outlier correspondence. Taking $\mathbf{x}_i$ as an example, consistent voting calculates candidate point via
\begin{equation}
    \mathbf y^l = NN_\mathbf{F}^l(\mathbf x_i, \mathbf{Y}), l = 1, 2, \cdots, L,
\end{equation}
where $NN_\mathbf{F}^l(\mathbf x, \mathbf{Y})$ reports the nearest neighbor of $\mathbf x$ in $\mathbf{Y}$ in feature $\mathbf{F}^l$'s space. We gradually check whether there exists two candidates ($\mathbf{y}^l, \mathbf{y}^{l+1}$) are the same from level $1$ to $L-1$, in which case 
there are at least two feature spaces helps to find the same candidate. So it is regard as a consistent correspondence and set $\sigma(i)$ to the index of $\mathbf y^l$ (the smallest $l$). Otherwise, the correspondence is considered to be unreliable and set $\sigma(i)$ to -1. A voting diagram for $L=3$ based on multi-level consistency is shown in \cref{fig:consistency}(c).

\begin{figure*}[!t]
\centering
\vspace{-.25em}
\includegraphics[width=.84\textwidth]{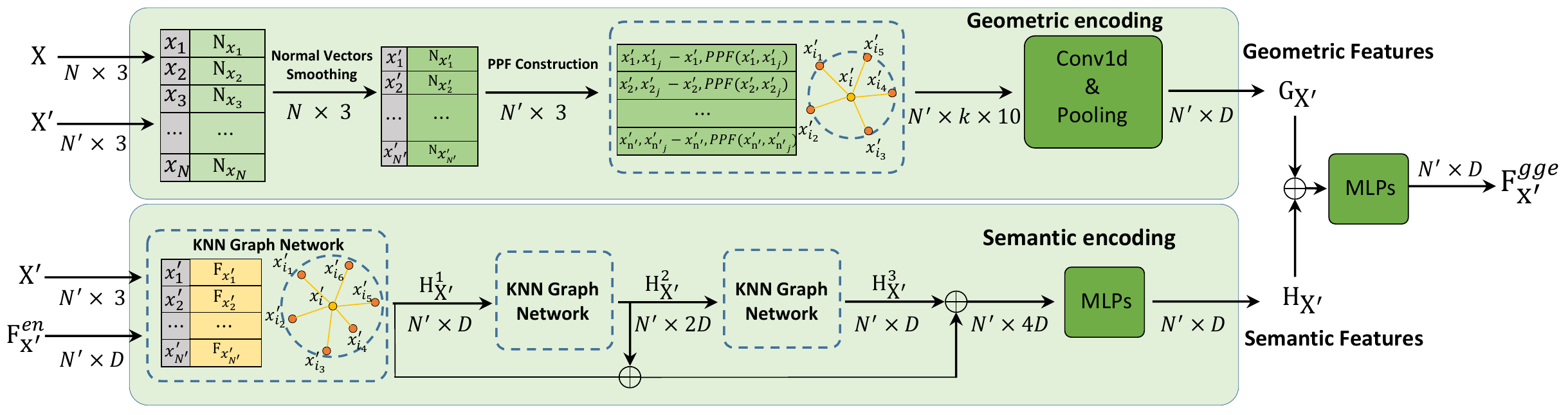}
\vspace{-.75em}
\caption{Geometry-guided encoding module}
\label{fig:gge}
\vspace{-1.5em}
\end{figure*}

\subsection{Geometry-guided encoding}
\label{subsec:gge}
Geometry-guided encoding (GGE) utilizes both geometric and semantic information to enhance the point feature and to recall more correct correspondences. However, it’s challenging to involve geometric encoding for real-world dataset in a one-forward-pass manner due to the large points number. Thus, GGE module is involved between the encoder and decoder, which takes the $\mathbf{X}'$ (the $L$-th downsampled $\mathbf{X}_L)$, $D$ dimensional $\mathbf{F}_\mathbf{X'}$ and original $\mathbf{X}$  as input, and outputs geometry enhanced features $\mathbf{F}_\mathbf{X'}^{gge}$ (\cref{fig:gge}).

\textbf{Normal vectors smoothing}. 
Normal vectors are used to enhance the geometric encoding\cite{rusu2008aligning,rusu2009fast}. Specifically, a smoothing strategy is proposed in GGE module to ensure the correct geometric presentation of the normals. Firstly, the subsampled superpoints $\mathbf{X}'$ are mapped back to the original $\mathbf{X}$. Then, the normals $\mathbf{N_{X}}$ for $\mathbf{X}$ are calculated with Open3D's classic API\cite{zhou2018open3d}. Finally, the normal vector of a certain point in  $\mathbf{X}'$ is calculated by averaging the normals of its surrounding points in $\mathbf{X}$. 

\textbf{Geometric encoding}. The geometric features ${\mathbf G}_{\mathbf{x}'_i}$ for point $\mathbf{x}'_i \in \mathbf{X}'$ is constructed with PPF\cite{drost2010model}, i.e., 
\begin{equation}
\begin{aligned}
\text{PPF}(\mathbf{x}_i', \mathbf{x}_j') =& (\angle(\mathbf{x}_{j}' - \mathbf{x}_i', \mathbf{N}_{\mathbf{x}'_i}), \angle (\mathbf{x}_{j}' - \mathbf{x}_i',\mathbf{N}_{\mathbf{x}'_j}),\\
&\angle (\mathbf{N}_{\mathbf{x}'_i}, \mathbf{N}_{\mathbf{x}'_j}), ||\mathbf{x}'_{i} - \mathbf{x}'_j||_2), \\ 
\mathbf{G}_{\mathbf{x}'_j} =& {f_1}(\mathbf{x}'_i, \mathbf{x}'_j - \mathbf{x}'_i, \text{PPF}(\mathbf{x}'_i, \mathbf{x}'_j)), \\
\mathbf{G}_{\mathbf{x}'_i} =& \max \{\mathbf{G}_{\mathbf{x}'_j}\,|\,\mathbf{x}'_j \in J^\mathbf{G}_i \}.
\end{aligned}
\vspace{-.5em}
\end{equation}
where $\angle(\cdot, \cdot) \in [0, \pi]$ denotes the angle between the two vectors, ${f_1}$ is implemented by PointNet\cite{qi2017pointnet}, $J^\mathbf{G}_i = \left\{\mathbf{x}'_j\,| \,||\mathbf{x}'_j - \mathbf{x}'_i|| < r^\mathbf{G}\right\}$, $r^\mathbf{G}$ is the radius of $\mathbf{x}'_i$'s neighborhood, 
and $\max(\cdot)$ means point-wise max-pooling.

\textbf{Semantic encoding.} Inspired by\cite{huang2021predator}, the dense connected graph network is introduced in GGE module to enhance the semantic features $\mathbf{F}_\mathbf{X'}^{en}$ (outputted from the encoder) as $\mathbf{H}_\mathbf{X'}$, which is constructed as follows,
\begin{equation}
\begin{aligned}
\mathbf{H}^{0}_{\mathbf{X}'} &= \mathbf{F}_{\mathbf{X}'}^{en},\, \mathbf{H}^i_{\mathbf{X}'} = \text{GNN}_i(\mathbf{H}^{i-1}_{\mathbf{X}'}), \\
\mathbf{H}_{\mathbf{X}'} &= \text{MLP}_1(\text{cat}(\mathbf{H}^1_{\mathbf{X}'}, \mathbf{H}^2_{\mathbf{X}'}, \mathbf{H}^3_{\mathbf{X}'})).
\end{aligned}
\end{equation}
Then we explicitly fuse the geometric features $\mathbf{G}_{\mathbf{X'}}$ with the semantic features $\mathbf{H}_{\mathbf{X'}}$ to generate the augmented $D$ dimensional features: 
\begin{equation}
\begin{aligned}
\mathbf{F}_\mathbf{X'}^{gge} = \text{MLP}_2(\text{cat}(\mathbf{G}_\mathbf{X'},\mathbf{H}_\mathbf{X'})).
\end{aligned}
\end{equation}

\begin{figure*}[!t]
\centering
\vspace{-.75em}
\includegraphics[width=.85\textwidth]{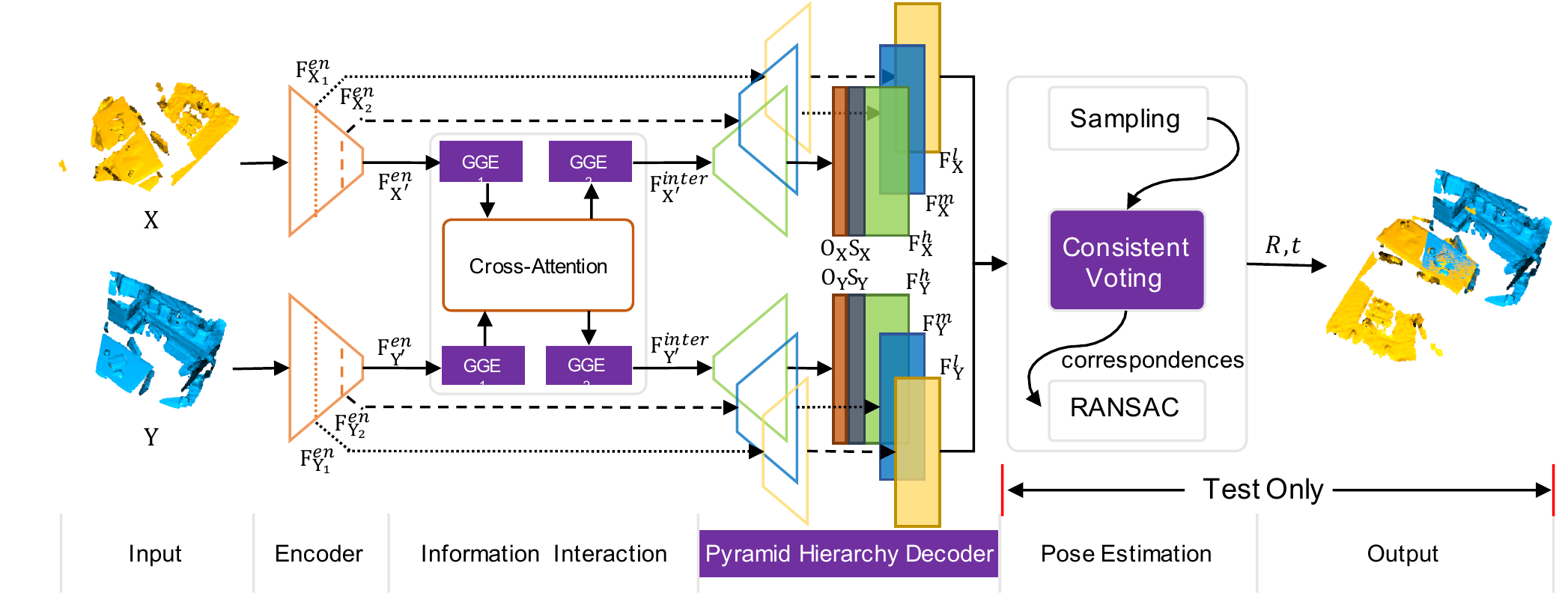}
\vspace{-1.25em}
\caption{The pipeline of GCNet for point cloud registration. Background color marked with purple are the proposed modules.}
\vspace{-1.5em}
\label{fig:gcnet}
\end{figure*}

\vspace{-1.em}
\subsection{Geometry-guided consistent network}
\vspace{-.25em}
Based on the proposed modules, we build our geometry-guided consistent network (GCNet). The full pipeline for point cloud registration is illustrated in \cref{fig:gcnet}.

\textbf{Encoder.} GCNet is built on KP-FCNN\cite{thomas2019kpconv} with sampling layers number $L$ being 3. For $\mathbf{X}$, the feature maps before each strided KPConv block are generated, denoted as $\mathbf{F}^{en}_{\mathbf{X}_1}, \mathbf{F}^{en}_{\mathbf{X}_2}, \mathbf{F}^{en}_\mathbf{X'} (\mathbf{F}^{en}_{\mathbf{X}_3})$. Encoder is shared by $\mathbf{X}$ and $\mathbf{Y}$.

\textbf{Information cross interaction.} The information cross interaction module bridges the encoder and decoder, and further translates the $\mathbf{F}_\mathbf{X'}^{en}$ and $\mathbf{F}_\mathbf{Y'}^{en}$ to information interacted features $\mathbf{F}_\mathbf{X'}^{inter}, \mathbf{F}_\mathbf{Y'}^{inter}$, which consists of two shared GGE modules and one cross attention module. 

The cross attention is a multi-head attention with residual connection.
For $\mathbf{X}'$ and $\mathbf{F}_\mathbf{X'}^{gge}$ ($\text{GGE}_1$'s output), the information interacted feature $\mathbf{F}_\mathbf{X'}^{inter}$ is calculated as 
\begin{equation}
\begin{aligned}
&\mathbf{Q}_i = \mathbf{F}_\mathbf{X'}^{gge} \cdot \mathbf{W}_i^\mathbf{Q} ,\, \mathbf{K}_i = \mathbf{F}_\mathbf{Y'}^{gge} \cdot \mathbf{W}_i^\mathbf{K}, \, \mathbf{V}_i = \mathbf{F}_\mathbf{Y'}^{gge} \cdot \mathbf{W}_i^\mathbf{V}\\
&\text{head}_i = \text{softmax}(\frac{\mathbf{Q}_i \cdot \mathbf{K}^T_i}{\sqrt{D_i}}) \cdot \mathbf{V}_i, \\
&\mathbf{F}^{ca}_{\mathbf{X}'} = \text{MLP}_2(\mathbf{F}_{\mathbf{X}'}^{gge} + \text{MLP}_1(\text{cat}(\text{head}_1, ..., \text{head}_k))), \\
&\mathbf{F}_\mathbf{X'}^{inter} = \text{GGE}_2(\mathbf{X}', \mathbf{F}^{ca}_{\mathbf{X}'}, \mathbf{X})
\end{aligned}
\hspace{-2.1em}
\end{equation}
where 
$\mathbf{W}_i^\mathbf{Q} \in \mathbb{R}^{D \times D_i}$, $\mathbf{W}_i^\mathbf{K} \in \mathbb{R}^{D \times D_i}$,  $\mathbf{W}_i^\mathbf{V} \in \mathbb{R}^{D \times D_i}$ are learnable weights matrices, $k$ is the number of heads, $D_i = D / k$, and $\text{GGE}_2(\cdot, \cdot, \cdot)$ is the GGE module defined in subsection \ref{subsec:gge}. 
$\mathbf{F}^{inter}_{\mathbf{Y}'}$ is obtained in the same way.

\textbf{Decoder.} The shared pyramid hierarchy decoder accepts $(\mathbf{F}^{en}_{\mathbf{X}_1}, \mathbf{F}^{en}_{\mathbf{X}_2}, \mathbf{F}^{inter}_{\mathbf{X'}})$ associated with the point coordinates as input, and generates three-level $C$ dimensional features $(\mathbf{F}_\mathbf{X}^h, \mathbf{F}_\mathbf{X}^m, \mathbf{F}_\mathbf{X}^l)$\footnote{Superscript is marked with $h$(high), $m$(middle), $l$(low) for clear representation.} for $\mathbf{X}$. Additionally, inspired by PREDATOR\cite{huang2021predator}, the overlap scores $\mathbf{O}_\mathbf{X} \in \mathbf{R}^{N \times 1}$ and saliency scores $\mathbf{S}_\mathbf{X} \in \mathbf{R}^{N \times 1}$ are calculated along with $\mathbf{F}_\mathbf{X}^h$, to provide the probability for each point in salient overlapping point sampling.


\textbf{Pose estimation.} Pose estimation starts with sampling a specific number(such as 5000) of points associated with the multi-level features based on $(\mathbf{O}_\mathbf{X}, \mathbf{S}_\mathbf{X})$ and $(\mathbf{O}_\mathbf{Y}, \mathbf{S}_\mathbf{Y})$ in probability. Then the consistent voting accepts the sampled $(\mathbf{X}, \mathbf{F}_\mathbf{X}^h, \mathbf{F}_\mathbf{X}^m, \mathbf{F}_\mathbf{X}^l)$ and $(\mathbf{Y}, \mathbf{F}_\mathbf{Y}^h, \mathbf{F}_\mathbf{Y}^m, \mathbf{F}_\mathbf{Y}^l)$ as input, and generate consistent correspondences. Finally, the transformation $(R, t)$ is estimated by RANSAC.

\subsection{Loss function for GCNet}

\textbf{Feature loss.} Circle loss for feature $\mathbf{F}_\mathbf{X}^h, \mathbf{F}_\mathbf{X}^m, \mathbf{F}_\mathbf{X}^l$ and $\mathbf{F}_\mathbf{Y}^h, \mathbf{F}_\mathbf{Y}^m, \mathbf{F}_\mathbf{Y}^l$ are calculated with the randomly sampled correspondences set $\mathcal{C}=\left\{(\mathbf{x}_i, \mathbf{y}_{\sigma(i)})\,|\,i = 1, 2, \cdots, S\right\}$, where $S$ is the number of sampled correspondences. For example, the loss $\mathcal L^h_\mathbf{X}(\mathbf{F})$ for $\mathbf{F}_\mathbf{X}^h$ is defined as
\begin{equation}
\begin{aligned}
\mathcal L^h_\mathbf{X}(\mathbf{F}) = \frac{1}{S}\sum_{i=1}^S\log[&1 + \sum_{\mathbf{y}_j \in \mathcal P_i} \exp(\gamma\alpha^p_{ij}(D^h_{ij} - \Delta p)) \\
&\cdot\sum_{\mathbf{y}_k \in \mathcal N_i} \exp(\gamma\alpha^n_{ik}(\Delta n - D^h_{ik}))],
\end{aligned}
\hspace{-2.5em}\vspace{-.5em}
\end{equation}
where $\mathcal P_i=\left\{\mathbf{y}_{\sigma(i)}\right\}$, $\mathcal N_i=\left\{\mathbf{y} \,|\, (\mathbf{x}, \mathbf{y}) \in \mathcal{C}, \mathbf{x} \ne \mathbf{x}_i \right\}$, $D_{ij}^h = ||\mathbf{F}^h_{\mathbf{x}_i} - \mathbf{F}^h_{\mathbf{y}_j}||_2$, $\Delta p$ and $\Delta n$ denote positive and negative margin, $\alpha_{ij}^p = [D_{ij}^h - \Delta p]_+$, $\alpha_{ik}^n = [\Delta n - D_{ik}^h]_+$, and $\gamma$ is a scale factor. The loss $\mathcal L^h_\mathbf{Y}(\mathbf{F})$ is defined in the same way. 
Then, let $\mathcal L^h(\mathbf{F}) = \frac{1}{2}(\mathcal{L}^h_{\mathbf{X}}(\mathbf{F}) + \mathcal{L}^h_{\mathbf{Y}}(\mathbf{F}))$, and so are the $\mathcal{L}^m(\mathbf{F})$ and $\mathcal{L}^l(\mathbf{F})$. 

\textbf{Overlap and saliency loss.} Binary cross entropy loss is adopted for $\mathbf{O}_\mathbf{X}$ and $\mathbf{S}_\mathbf{X}$, i.e.,
\begin{equation}
\begin{aligned}
    \mathcal{L}_\mathbf{X}(\mathbf{O}) &= -\sum_{\mathbf{x}_i\in\mathbf{X}} w_i^\mathbf{O}  [\overline{\mathbf{O}}_{\mathbf{x}_i}  \log \mathbf{O}_{\mathbf{x}_i}
    + (1 - \overline{\mathbf{O}}_{\mathbf{x}_i}) \log( 1 - \mathbf{O}_{\mathbf{x}_i})], \\
    \mathcal{L}_\mathbf{X}(\mathbf{S}) &= -\sum_{\mathbf{x}_i\in\mathbf{X}} w_i^\mathbf{S} [\overline{\mathbf{S}}_{\mathbf{x}_i}  \log \mathbf{S}_{\mathbf{x}_i}
    + (1 - \overline{\mathbf{S}}_{\mathbf{x}_i})  \log( 1 - \mathbf{S}_{\mathbf{x}_i})],
\end{aligned}\vspace{-.5em}
\end{equation}
where $w_i^\mathbf{O}$ and $w_i^\mathbf{S}$ are weighting factors for category balance, $\overline{\mathbf{O}}_{\mathbf{x}_i}$ and $\overline{\mathbf{S}}_{\mathbf{x}_i}\in\{0,1\}$ are the ground truth binary overlap score and saliency score. $\overline{\mathbf{O}}_{\mathbf{x}_i}$ is set to 
1 if the distance between $T^*(\mathbf{x}_i)$ and $NN(T^*(\mathbf{x}_i), \mathbf{Y})$ is below the threshold, where $T^*$ is the ground truth transformation and $NN(\mathbf{x}, \mathbf{Y})$ operator reports the nearest neighbor of $\mathbf{x}$ in $\mathbf{Y}$. $\overline{\mathbf{S}}_{\mathbf{x}_i}$ is set to 1 if the ground truth candidate point is matched for $\mathbf{x}_i$ based on feature matching. $\mathcal{L}_\mathbf{Y}(\mathbf{O})$ and $\mathcal{L}_\mathbf{Y}(\mathbf{S})$ are defined in the same way. Then, let 
$\mathcal L(\mathbf{O}) = \frac{1}{2}(\mathcal{L}_{\mathbf{X}}(\mathbf{O}) + \mathcal{L}_{\mathbf{Y}}(\mathbf{O})), \mathcal L(\mathbf{S}) = \frac{1}{2}(\mathcal{L}_{\mathbf{X}}(\mathbf{S}) + \mathcal{L}_{\mathbf{Y}}(\mathbf{S}))$. 

\textbf{Combined loss.} The complete loss function is 
\begin{equation}
\mathcal{L} = \mathcal{L}^h(\mathbf{F}) + \mathcal{L}^m(\mathbf{F}) + \mathcal{L}^l(\mathbf{F}) + \mathcal{L}(\mathbf{O}) + \mathcal{L}(\mathbf{S}).
\end{equation}

\vspace{-1.75em}
\section{Experiments}
\vspace{-.25em}
We first evaluated and analysed the proposed geometry-guided consistent (GC) mechanisms on learning-based and traditional registration methods across two challenging datasets (3DMatch\cite{zeng20173dmatch} and 3DLoMatch\cite{huang2021predator}). Then GCNet was evaluated on 3DMatch\cite{zeng20173dmatch}, 3DLoMatch\cite{huang2021predator}, Odometry KITTI\cite{geiger2012we} and MVP-RG\cite{pan2021robust} datasets, resulting in a detailed quantitative analysis. 

\vspace{-.25em}
\subsection{Implementation details}
\vspace{-.25em}
Three learning-based models (FCGF\cite{choy2019fully}, D3Feat\cite{bai2020d3feat}, PREDATOR\cite{huang2021predator}) and one traditional hand-crafted FPFH\cite{rusu2009fast} were involved in our experiments to evaluate the GC mechanisms. For the learning-based models, we retrained the source code provided by the authors by adding the GC mechanisms. For FPFH which doesn't require training, only pyramid hierarchy decoder and consistent voting were considered. The multi-scale features $\mathbf{F}^h, \mathbf{F}^m$ and $\mathbf{F}^l$ were implemented with decreasing radius(15, 10 and 5 times of voxel size). GCNet was implemented in PyTorch\cite{paszke2019pytorch}, with knn parameter $k$ of PPF set to 64, the dimension $D$ of $\mathbf{F}^{gge}$ set to 256 and the dimension $C$ of $\mathbf{F}^{fin}$ set to 32. Parameter $d$ in voting was set to $2\times$ voxel size. The number of sampled point pairs $K$ set to 256, 512 and 768 for 3DMatch (3DLoMatch), Odometry KITTI and MVP-RG, respectively. The scale factor $\gamma$, positive and negative margin $\Delta p, \Delta n$ in loss function were set to 16, 0.1 and 1.4.


\begin{table*}[!h]
\centering
\small
\vspace{-.25em}
\scalebox{.88}{
\begin{tabular}{l|ll|ll|cc}
\hline
& \multicolumn{2}{c}{3DMatch} & \multicolumn{2}{c}{3DLoMatch} &  \\
Model & IR (\%) $\uparrow$ & RR (\%) $\uparrow$ & IR (\%) $\uparrow$ & RR (\%) $\uparrow$ & \# Params $\downarrow$ & time (ms) $\downarrow$  \\
\hline
FPFH\cite{rusu2009fast} & 11.2 & 62.3 & 3.2 & 20.5 & {\bf 0} & {\bf 373} \\
FPFH + GC & {\bf 18.2 (+7.0)} & {\bf 73.7 (+11.4)} & {\bf 5.7 (+2.5)} & {\bf 31.4 (+10.9)} & {\bf 0} & 648 \\
FCGF\cite{choy2019fully} & 34.1 & 85.0 & 10.0 & 51.3 & {\bf 8.76M} & 6803 \\
FCGF + GC & {\bf 48.4 (+14.3)} & {\bf 89.7 (+4.7)} & {\bf 16.6 (+6.6)} & {\bf 59.9 (+8.6)} & 10.64M & {\bf 4343} \\
D3Feat\cite{bai2020d3feat} & 22.6 & 83.2 & 5.0 & 36.2 & {\bf 24.3M} & 141 \\
D3Feat + GC & {\bf 35.7 (+13.1)} & {\bf 87.0 (+4.8)} & {\bf 10.0 (+5.0)} & {\bf 47.9 (+11.7)} & 37.19M & {\bf 123} \\
PREDATOR\cite{huang2021predator} & 43.8 & 89.0 & 16.8 & 57.2 & {\bf 7.43M} & {\bf 797}  \\
PREDATOR + GC & {\bf 59.8 (+16.0)} & {\bf 90.7 (+1.7)} & {\bf 31.0 (+14.2)} & {\bf 66.9 (+9.7)} & 8.86M & 864\\
\hline
\end{tabular}}
\vspace{-.75em}
\caption{Results of transferring GC mechanisms to other registration methods. IR without mutually feature matching were reported and improvements are shown in brackets. Average time was tested on 3DLoMatch.}
\label{tab:flex}
\end{table*}

\begin{figure*}[!t]
\centering
\vspace{-1.25em}
\centerline{\includegraphics[width=0.85\textwidth]{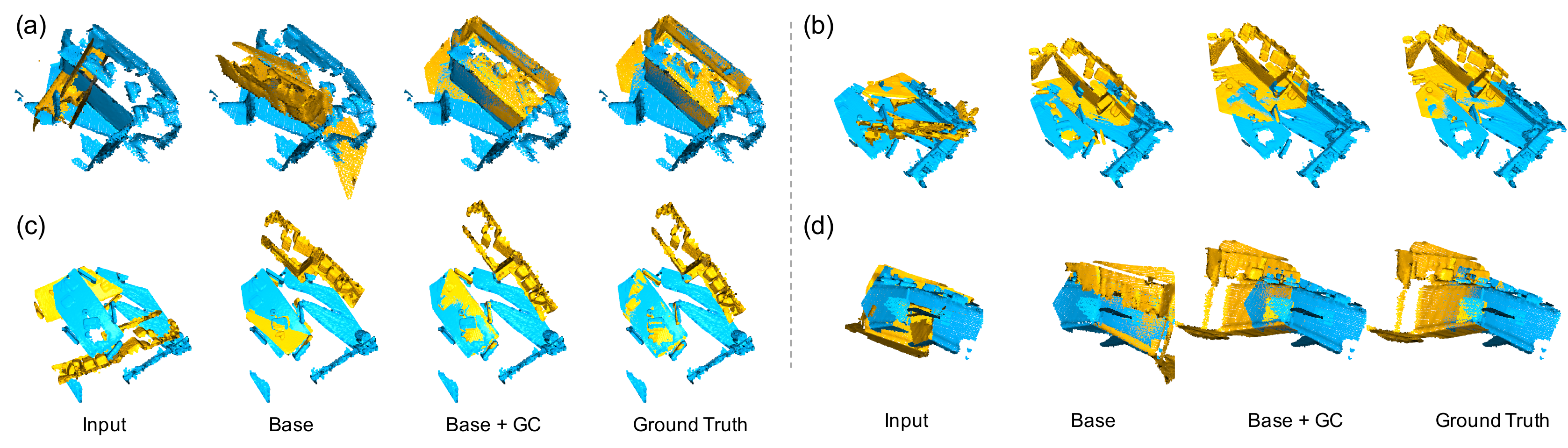}}
\vspace{-1em}
\caption{Improvements gained by GC mechanism. Base denotes FPFH, FCGF, D3Feat and PREDATOR in (a)(b)(c)(d), respectively.}
\vspace{-1.75em}
\centering
\label{fig:base_gc}
\end{figure*}

\vspace{-.25em}
\subsection{3DMatch and 3DLoMatch}
\vspace{-.25em}
3DMatch\cite{zeng20173dmatch} and 3DLoMatch\cite{huang2021predator} are two widely used indoor datasets that contain \textgreater $\large30\%$ and $10\sim30\%$ partial overlapping scene pairs, respectively.
In consistent with \cite{huang2021predator}, 46 scenes (20642 scan pairs) were used for training and 8 scenes (1331 scan pairs) were used for validation. 8 scenes (1279 and 1726 non-consecutive pairs for 3DMatch and 3DLoMatch) were used for testing. 
The {\em Inlier Ratio} (IR), {\em Feature Matching Recall} (FMR), {\em Registration Recall} (RR), {\em Relative Translation Error} (RTE) and {\em Relative Rotation Error} (RRE) for each dataset were reported (detailed definition of these criteria can be found in Supplement S1).

\vspace{-1.25em}
\subsubsection{Transfer GC mechanisms to other methods}
\vspace{-.25em}
The flexibility and effectiveness of the proposed GC mechanisms were validated by transferring them to other feature-based methods. Experiments have been conducted on the 3DMatch and 3DLoMatch, and the results are shown in \cref{tab:flex}. Here, it can be found that GC mechanisms greatly improve the baseline methods' performance.  Besides, using GC mechanisms achieve less or on par running time compared with the learning-based baselines. One reason is that the encoder is shared by the pyramid hierarchy decoder. More importantly, many points with non-robust features are abandoned via consistent voting, which decreases nearest search in RANSAC iterations. For model size, slightly parameter number increases (about 1.2$\sim$1.5x papameters) compared with the baselines. Several registration cases are visualized in \cref{fig:base_gc}. As observed, the base methods work well with GC mechanisms.

\begin{table}[b!]
\setlength{\tabcolsep}{2pt}
\centering\scriptsize
\vspace{-2.5em}
\begin{tabular}{lccccc|ccccc}
\toprule
& \multicolumn{5}{c}{3DMatch} & \multicolumn{5}{c}{3DLoMatch} \\
\midrule
\# Samples  & 5000 & 2500 & 1000 & 500 & 250 & 5000 & 2500 & 1000 & 500 & 250 \\
\midrule
& \multicolumn{10}{c}{{\it Registration Recall (\%)} $\uparrow$} \\
\hline
3DSN\cite{gojcic2019perfect} & 78.4 & 76.2 & 71.4 & 67.6 & 50.8 & 33.0 & 29.0 & 23.3 & 17.0 & 11.0 \\
FCGF\cite{choy2019fully} & 85.1 & 84.7 & 83.3 & 81.6 & 71.4 & 40.1 & 41.7 & 38.2 & 35.4 & 26.8 \\
D3Feat\cite{bai2020d3feat} & 81.6 & 84.5 & 83.4 & 82.4 & 77.9 & 37.2 & 42.7 & 46.9 & 43.8 & 39.1 \\
PREDATOR\cite{huang2021predator} & 89.0 & 89.9 & 90.6 & 88.5 & 86.6 & 59.8 & 61.2 & 62.4 & 60.8 & 58.1 \\
CoFiNet\cite{yu2021cofinet} & 89.3 & 88.9 & 88.4 & 87.4 & 87.0 & 67.5 & 66.2 & 64.2 & 63.1 & 61.0 \\
RegTR\cite{yew2022regtr} & \underline{92.0} & - & - & - & - & 64.8 & - & - & - & - \\
GeoTransformer\cite{qin2022geometric} & \underline{92.0} & \underline{91.8} & \underline{91.8} & \textbf{91.4} & \textbf{91.2} & \textbf{75.0} & \textbf{74.8} & \textbf{74.2} & \textbf{74.1} & \textbf{73.5} \\
GCNet ({\em ours}) & \textbf{92.9} & \textbf{92.1} & \textbf{92.3} & \underline{90.3} & \underline{88.7} & \underline{71.9} & \underline{71.4} & \underline{70.6} & \underline{67.5} & \underline{61.9} \\
\midrule
& \multicolumn{10}{c}{{\it Feature Matching Recall (\%)} $\uparrow$} \\
\midrule
3DSN\cite{gojcic2019perfect} & 95.0 & 94.3 & 92.9 & 90.1 & 82.9 & 63.6 & 61.7 & 53.6 & 45.2 & 34.2 \\
FCGF\cite{choy2019fully} & 97.4 & 97.3 & 97.0 & 96.7 & 96.6 & 76.6 & 75.4 & 74.2 & 71.7 & 67.3 \\
D3Feat\cite{bai2020d3feat} & 95.6 & 95.4 & 94.5 & 94.1 & 93.1 & 67.3 & 66.7 & 67.0 & 66.7 & 66.5 \\
PREDATOR\cite{huang2021predator} & 96.6 & 96.6 & 96.5 & 96.3 & 96.5 & 78.6 & 77.4 & 76.3 & 75.7 & 75.3 \\
CoFiNet\cite{yu2021cofinet} & \underline{98.1} & \underline{98.3} & \textbf{98.1} & \underline{98.2} & \textbf{98.3} & 83.1 & 83.5 & 83.3 & 83.1 & 82.6 \\
RegTR\cite{yew2022regtr} & - & - & - & - & - & - & - & - & - & - \\
GeoTransformer\cite{qin2022geometric} & 97.9 & 97.9 & \underline{97.9} & 97.9 & 97.6 & \textbf{88.3} & \textbf{88.6} & \textbf{88.8} & \textbf{88.6} & \textbf{88.3}\\
GCNet ({\em ours}) & \textbf{98.2} & \textbf{98.4} & \textbf{98.1} & \textbf{98.6} & \underline{98.2} & \underline{84.5} & \underline{84.8} & \underline{85.0} & \underline{85.3} & \underline{83.4} \\
\midrule
& \multicolumn{10}{c}{{\it Inlier Ratio (\%)} $\uparrow$} \\
\midrule
3DSN\cite{gojcic2019perfect} & 36.0 & 32.5 & 26.4 & 21.5 & 16.4 & 11.4 & 10.1 & 8.0 & 6.4 & 4.8 \\
FCGF\cite{choy2019fully} & 56.8 & 54.1 & 48.7 & 42.5 & 34.1 & 21.4 & 20.0 & 17.2 & 14.8 & 11.6 \\
D3Feat\cite{bai2020d3feat} & 39.0 & 38.8 & 40.4 & 41.5 & 41.8 & 13.2 & 13.1 & 14.0 & 14.6 & 15.0 \\
PREDATOR\cite{huang2021predator} & 58.0 & 58.4 & 57.1 & 54.1 & 49.3 & 26.7 & 28.1 & 28.3 & 27.5 & 25.8 \\
CoFiNet\cite{yu2021cofinet} & 49.8 & 51.2 & 51.9 & 52.2 & 52.2 & 24.4 & 25.9 & 26.7 & 26.8 & 26.9  \\
RegTR\cite{yew2022regtr} & - & - & - & - & - & - & - & - & - & - \\
GeoTransformer\cite{qin2022geometric} & \textbf{71.9} & \textbf{75.2} & \textbf{76.0} & \textbf{82.2} & \textbf{85.1} & \textbf{43.5} & \textbf{45.3} & \textbf{46.2} & \textbf{52.9} & \textbf{57.7} \\
GCNet ({\em ours}) & \underline{63.1} & \underline{63.5} & \underline{61.5} & \underline{57.6} & \underline{51.1} & \underline{31.8} & \underline{33.3} & \underline{33.5} & \underline{31.9} & \underline{29.2} \\
\bottomrule
\end{tabular}
\vspace{-1.25em}
\caption{Results on 3DMatch and 3DLoMatch under different numbers of sampling points.}
\vspace{-.5em}
\label{tab:3dmatch_3dlomatch}
\end{table}

\vspace{-1.75em}
\subsubsection{GCNet vs. the state-of-the-art methods}
\vspace{-.5em}


\begin{figure*}[!t]
\centering
\centerline{\includegraphics[width=0.93\textwidth]{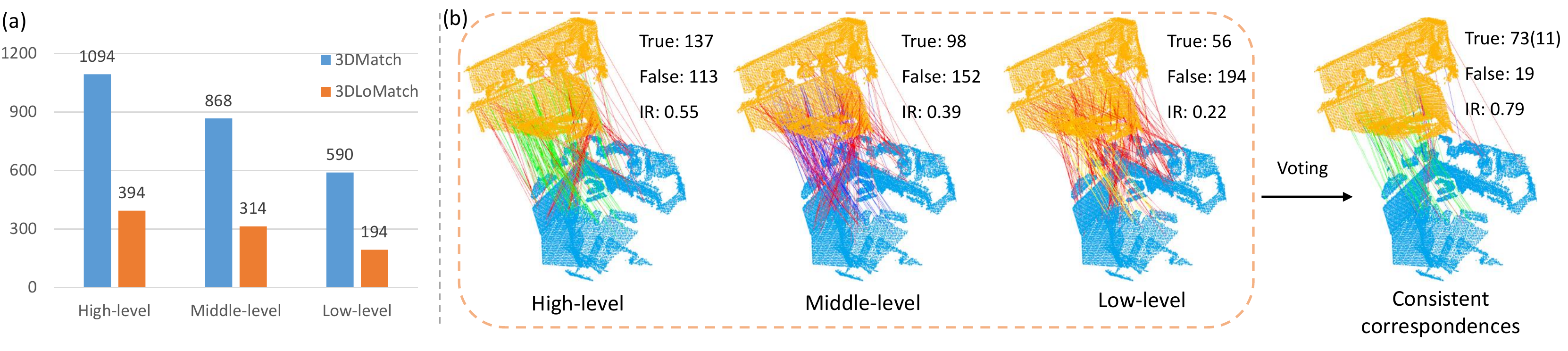}}
\vspace{-1em}
\caption{
(a) Inlier points number on 3DMatch and 3DLoMatch through different-level features matching. (b) Correspondences constructed by multi-level feature matching and consistent voting.
250 correspondences are randomly selected for convenience of visualization. Here, the correct correspondences obtained based on high-, middle- and low-level features are marked in green, blue and yellow, respectively. The incorrect correspondences are marked in red. 
The correspondences constructed contain 11 points with middle-level features and 62 points with high-level features, having preserved most of the correct ones and rejected most of the specious features.}
\vspace{-1.75em}
\label{fig:phd-voting}
\end{figure*}

GCNet was compared with the SOTA method 3DSN\cite{gojcic2019perfect}, FCGF\cite{choy2019fully}, D3Feat\cite{bai2020d3feat}, PREDATOR\cite{huang2021predator}, CoFiNet\cite{yu2021cofinet}, RegTR\cite{yew2022regtr} and GeoTransformer\cite{qin2022geometric} on 3DMatch and 3DLoMatch datasets (results shown in \cref{tab:3dmatch_3dlomatch}) and outperforms all the other models on 3DMatch with 92.9\% RR.
Additionally, GCNet performs well under various number of sampling points, demonstrating the robustness of the model. GCNet achieves the second best performance on 3DLoMatch dataset with 71.9\% RR. When 250 points are sampled, the performance of GCNet drops a lot on 3DLoMatch. GeoTransformer performs well due to it was trained on the downsampled superpoints. We didn't pay more attention on such situation for it is too sparse for the low overlap pairs which may not perform well stably in real-world applications. FMR and IR are also reported. GeoTransformer achieves higher IR under a local-to-global two-step registration scheme. However, it is reasonable because the others just make direct matching, while GCNet achieves the best among these methods.   
Several difficult cases can be found in Supplement Figure S1. 

\begin{table}[t!]
\setlength{\tabcolsep}{4pt}
\centering\footnotesize
\vspace{.25em}
\begin{tabular}{cccc|ccc}
\toprule
BASE & PHD & CV & GGE & IR (\%) $\uparrow$ & FMR (\%) $\uparrow$ & RR (\%) $\uparrow$ \\
\midrule
$\checkmark$ & & & & 50.3 & 96.3 & 89.3 \\
$\checkmark$ & $\checkmark$ & & & 56.6 & 97.2 & 90.3 \\
$\checkmark$ & $\checkmark$ & $\checkmark$ & & 60.8 & 98.1 & 91.5 \\
$\checkmark$ & & & $\checkmark$ & 54.4 & 97.4 & 91.2 \\
$\checkmark$ & $\checkmark$ & $\checkmark$ & $\checkmark$ & {\bf 63.1} & {\bf 98.2} & {\bf 92.9} \\
\bottomrule
\end{tabular}
\vspace{-.75em}
\caption{Ablation study on 3DMatch.}
\vspace{-2em}
\label{tab:ab_study}
\end{table}

\vspace{-1.25em}
\subsubsection{Scene performance} 
\vspace{-.5em}
We report {\em Registration Recall} (RR) on 3DMatch and 3DLoMatch in scene level, with additional evaluation metrics including  {\em Relative Rotation Error} (RRE) and {\em Relative Translation Error} (RTE). Limited by the space, the detailed results are shown in Supplement Table S1. 
Successfully registered pairs in each scene are included to calculate RRE and RTE. We achieve the highest RR in most scenes and the second highest in few scenes. Significant improvement has been made in RR, which indicates taking more difficult pairs into account. And there is a slight growth in RRE and RTE. It shows GCNet is robust and accurate for registration.

\vspace{-1.25em}
\subsubsection{Ablation study and components analysis}
\vspace{-.25em}

Pyramid hierarchy decoder (PHD), consistent voting (CV) and GGE modules are the three cornerstones to lift the inlier correspondences ratio and boost point cloud registration performance. Here, ablation study was conducted on 3DMatch dataset under 5000 sampling points, where the GCNet without PHD, CV and GGE modules was considered as the base model. \cref{tab:ab_study} summarizes the results of the ablation study, from which we can draw the conclusion that phd+voting promotes RR by 2.2\%, GGE promotes RR by 1.9\% and phd+voting+GGE promotes RR by 3.6\%. To our surprise, training the network with pyramid hierarchy decoder but only testing with high-level features (BASE+PHD) also improves the RR by 1.0\%.

\begin{figure}[!t]
\centering
\vspace{-.25em}
\centerline{\includegraphics[width=.86\linewidth]{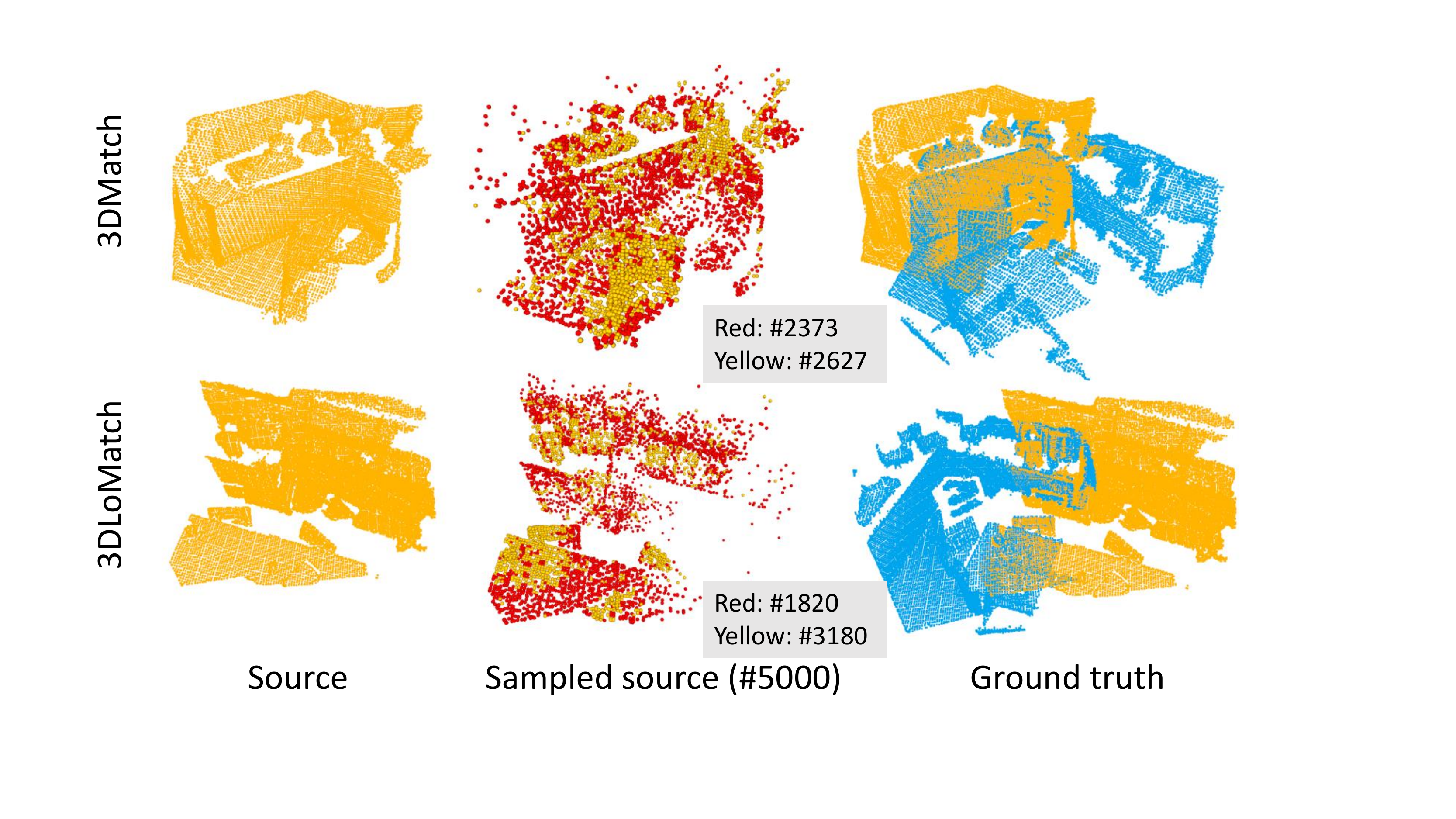}}
\vspace{-.75em}
\caption{
(a) Non-consistent points visualization. Consistent and non-consistent points are visualized in sampled source point cloud, which are marked in red and green. Non-consistent points are generally located in plane regions or some non-overlap regions.}
\vspace{-2em}
\label{fig:rm_vis}
\end{figure}

\textbf{Features of pyramid hierarchy decoder.} Pyramid hierarchy decoder generates high-, middle- and low-level features. We analysed them via calculating inlier points number recalled in their respective feature space. 
\cref{fig:phd-voting}(a) demonstrates that all the multi-level features are helpful for registration while high-level feature still performs best.

\textbf{Effects of consistent voting.} \cref{fig:phd-voting}(b) shows an example of how consistent voting works. On one hand, consistent voting prunes massive mismatched correspondences. On the other hand, it constructs correspondences via different-level features (green and blue lines), thus further lift the inlier correspondences proportion.  Non-consistent points abandoned by consistent voting distribution are also visualized on 3DMatch and 3DLoMatch in \cref{fig:rm_vis}.

\begin{table}[b!]
\setlength{\tabcolsep}{6pt}
\centering\footnotesize
\vspace{-2.05em}
\begin{tabular}{l|cccc}
\toprule
Points & \# points & $k$  & $D_p$ & Memory (MB) \\
\midrule
original points & 37001 (avg) & 1024 & 64 & 9,250 \\
\emph{superpoints} & 679 (avg) & 64 & 64 & 10 \\
\midrule
original points & 60000 (max) & 1024 & 64 & 15,000 \\
\emph{superpoints} & 1389 (max) & 64 & 64 & 22 \\
\midrule
original points & 37001 (avg) & 1024 & 256 & 37001 \\
\emph{superpoints} & 679 (avg) & 64 & 256 & 40 \\
\midrule
original points & 60000 (max) & 1024 & 256 & 60,000 \\
\emph{superpoints} & 1389 (max) & 64 & 256 & 43 \\
\bottomrule
\end{tabular}
\vspace{-.75em}
\caption{GPU memory consumption of PPF on 3DMatch.}
\label{tab:ppf_gpu}
\vspace{-.25em}
\end{table}

\textbf{PPF memory analysis.} We compared the GPU memory consumption of PPF on the original points and the downsampled \emph{superpoints} on 3DMatch train set. We first calculated the average and maximum points number of the original points and \emph{superpoints}. Then, following PPFNet\cite{deng2018ppfnet}, neighborhood points number $k$ under the specific radius was set to 1024 for original points, while $k$ was set to 64 in GCNet for \emph{superpoints}. The channel number $D_p$ for PPF was set to 64 and 256 for low- and high-dimension encoding, respectively. The GPU memory consumption for the feature map (without considering gradient map
) is shown in \cref{tab:ppf_gpu}, in which PPF's GPU memory consumption on \emph{superpoints} is three orders of magnitude smaller than on original points.

\subsubsection{Time analysis} 
\vspace{-.25em}
The runtime of pyramid hierarchy decoder (PHD), consistent voting (CV) and GGE modules was evaluated on 3DLoMatch. Three stages including feature extraction, RANSAC and the whole process were calculated separately. The results are shown in \cref{tab:time}. PHD+CV increases the feature extraction time from 57 to 78ms. However, less points(1412) in source point cloud (shown in \cref{fig:rm_vis}) are taken into consideration for RANSAC, thus the RANSAC runtime are reduced from 147 to 83ms. On the whole, it reduces the total time from 215ms to 181ms averagely. GGE module also increases the feature extraction time from 57ms to 95ms. Overall, equipped both PHD+CV and GGE module, GCNet achieves an average runtime of 226ms, which is comparable to the one of the base model (215ms).

\begin{table}[h!]
\setlength{\tabcolsep}{1pt}
\centering\footnotesize
\vspace{-.75em}
\begin{tabular}{l|cccc}
\toprule
Model & \# Points & Feature (ms) & RANSAC(ms) & Total (ms) \\
\midrule
BASE & 5000 & 57 & 147 & 215 \\
BASE + PHD + CV & 1412 & 78 & 83 & 181 \\
BASE + GGE & 5000 & 95 & 165 & 272 \\
GCNet & 1506 & 116 & 90 & 226 \\
\bottomrule
\end{tabular}
\vspace{-.5em}
\caption{Time analysis on 3DLoMatch.}
\vspace{-1.25em}
\label{tab:time}
\end{table}

\subsection{Odometry KITTI}


Odometry KITTI\cite{geiger2012we} is an outdoor LIDAR dataset for autonomous driving. In consistent with \cite{huang2021predator}, the frame pair at most 10 meters were selected to form scan pairs. 
Then, the model was trained with sequence No.00-05 (1358 scan pairs), validated with sequence No.06-07 (180 scan pairs) and tested with sequence No.08-10 (555 pairs). Finally, the {\em Relative Translation Error} (RTE), {\em Relative Rotation Error} (RRE) and {\em Registration Recall} (RR) of the results are reported (detailed definition of Registration Recall can be found in Supplement S4). 

\begin{table}[t]
\setlength{\tabcolsep}{1pt}
\centering\footnotesize
\begin{tabular}{l|ccc}
\toprule
Method & RTE (cm) $\downarrow$ & RRE ($^\circ$) $\downarrow$ & RR (\%) $\uparrow$ \\
\midrule
3DFeat-Net\cite{yew20183dfeat} & 25.9 & 0.57 & 96.0 \\
FCGF\cite{choy2019fully} & 9.5 & 0.30 & 96.6 \\
D3Feat\cite{bai2020d3feat} & 7.2 & 0.30 & {\bf 99.8} \\
PREDATOR\cite{huang2021predator} & 6.8 & 0.27 & {\bf 99.8} \\
CoFiNet\cite{yu2021cofinet} & 8.5 & 0.41 & {\bf 99.8} \\
HRegNet\cite{lu2021hregnet} & 12 & 0.29 & \underline{99.7} \\
GeoTransformer\cite{qin2022geometric} (RANSAC) & 7.4 & 0.27 & {\bf 99.8} \\
GeoTransformer\cite{qin2022geometric} (LGR) & 6.8 & {\bf 0.24} & {\bf 99.8} \\
\midrule
GCNet (wo. GGF) ({\em ours}) & \underline{6.3} & \underline{0.25} & {\bf 99.8} \\
GCNet ({\em ours}) & {\bf 6.1} & 0.26 & {\bf 99.8} \\
\bottomrule
\end{tabular}
\vspace{-.5em}
\caption{Results on Odometry KITTI dataset.}
\vspace{-1.75em}
\label{tab:kitti}
\end{table}

\vspace{-.5em}
\subsubsection{GCNet vs. the state-of-the-art methods}
\vspace{-.25em}
GCNet was compared with the SOTA method 3DFeat-Net\cite{yew20183dfeat}, FCGF\cite{choy2019fully}, D3Feat\cite{bai2020d3feat}, PREDATOR\cite{huang2021predator}, CoFiNet\cite{yu2021cofinet}, HRegNet\cite{lu2021hregnet} and GeoTransformer\cite{qin2022geometric} whose results are summarized in \cref{tab:kitti}. Here, GCNet achieve the best RR (99.8\%), RTE ($6.1 cm$) and the second best RRE ($0.26^\circ$). Limited by space, A case visualization on Odeometry KITTI can be seen in Supplement Figure S2. 
It should be noted that GCNet can achieve better RRE with $0.25^\circ$ without GGE module. Maybe normal vectors are not suitable for LIDAR dataset, which has relative weak structure information. 

\subsection{MVP-RG}

MVP(multi-view partial)-RG dataset\cite{pan2021variational,pan2021robust} is constructed by 16 categories object-centric synthetic point cloud. 
The density and unrestricted rotations ([0$^\circ$, 360$^\circ$]) vary in different point cloud. We used the official training set (6400 pairs) for training and used the official test set (1200 pairs) for testing.
Following \cite{pan2021robust}, we report three metrics $\mathcal{L}_R, \mathcal{L}_t$, and $\mathcal{L}_{RMSE}$ (detailed definition of these criteria can be found in Supplement S5).

\vspace{-.75em}
\subsubsection{GCNet vs. the state-of-the-art methods}
\vspace{-.25em}
GCNet was compared with PREDATOR\cite{huang2021predator}, and the recent end-to-end learning-based registration methods DCP\cite{wang2019deep}, RPM-Net\cite{yew2020rpm}, RGM\cite{fu2021robust} and GMCNet\cite{pan2021robust}. For PREDATOR and GCNet, both 768 correspondences were sampled for RANSAC. \cref{tab:mvp_rg} summarizes the results, where NgeNet outperforms all the other methods. Particularly, GCNet surpasses the end-to-end learning method in a remarkable margin. A case study can be found in Supplement Figure S2, which indicates that the consistent voting lift the inlier correspondences proportion effectively.
. 

\begin{table}[h!]
\setlength{\tabcolsep}{7pt}
\centering\small
\vspace{-.5em}
\begin{tabular}{l|ccc}
\toprule
Method & $\mathcal{L}_R$ ($^\circ$) $\downarrow$  & $\mathcal{L}_t \downarrow$ & $\mathcal{L}_{RMSE} \downarrow$ \\
\midrule
DCP\cite{wang2019deep} & 30.73 & 0.273 & 0.634 \\
RPM-Net\cite{yew2020rpm} & 22.20 & 0.174 & 0.327 \\
RGM\cite{fu2021robust} & 41.27 & 0.425 & 0.583 \\
GMCNet\cite{pan2021robust} & 16.57 & 0.174 & 0.246 \\
\midrule
PREDAOTR\cite{huang2021predator} & \underline{10. 58} & \underline{0.067} & \underline{0.125} \\
\midrule
GCNet ({\em ours}) & {\bf 7.99} & {\bf 0.048} & {\bf 0.093} \\
\bottomrule
\end{tabular}
\vspace{-.5em}
\caption{Results on MVP-RG.}
\vspace{-1.25em}
\label{tab:mvp_rg}
\end{table}

\vspace{-.5em}
\section{Conclusion and discussion}

In this paper, a pyramid hierarchy decoder coupled with consistent voting and geometry-guided encoding module were proposed to lift inlier correspondences proportion. The proposed techniques can be transferred to other learning-based and traditional methods to expediently boost the baseline performance. Additionally, we proposed GCNet for accurate point cloud registration, which outperforms state-of-the-art methods across multiple datasets.

Though the performance improvement gained by our proposed techniques, some failed registration still can not be avoid, which is demonstrated in Supplement Figure S3. 
Also, an end-to-end registration networks based on GC mechanisms will be further explored.

{\small
\bibliographystyle{ieee_fullname}
\bibliography{egbib}
}

\end{document}